\pdfoutput=1
\documentclass[11pt]{article}

\usepackage[]{ACL2023}

\usepackage{times}
\usepackage{latexsym}
\usepackage{graphicx}
\usepackage{amsmath}
\usepackage{booktabs}
\usepackage{arydshln}
\usepackage{bbm}
\usepackage{multirow}
\usepackage{makecell}
\usepackage{fdsymbol}
\usepackage{verbatim}

\usepackage{pifont}
\newcommand{\cmark}{\ding{51}}%
\newcommand{\xmark}{\ding{55}}%

\usepackage[T1]{fontenc}
\usepackage[utf8]{inputenc}

\usepackage{microtype}

\usepackage{inconsolata}

\title{ConcEPT: Concept-Enhanced Pre-Training for Language Models}

\author{
Xintao Wang$^1$, 
Zhouhong Gu$^1$, 
Jiaqing Liang$^1$\footnote{Corresponding Authors},  
~Dakuan Lu$^1$, 
Yanghua Xiao$^{1,2*}$, 
Wei Wang$^1$
\\
$^1$Shanghai Key Laboratory of Data Science, School of Computer Science, Fudan University\\
$^2$Fudan-Aishu Cognitive Intelligence Joint Research Center\\
xtwang21@m.fudan.edu.cn,
{l.j.q.light, ludakuan1234}@gmail.com,
{zhgu20, shawyh, weiwang1}@fudan.edu.cn
}

\begin{document}
\maketitle

\let\thefootnote\relax\footnote{*Corresponding Authors.}
\let\thefootnote\relax\footnote{Work completed in 2023.01.}

\begin{abstract}
Pre-trained language models (PLMs) have been prevailing in state-of-the-art methods for natural language processing, and
knowledge-enhanced PLMs are further proposed to promote model performance in knowledge-intensive tasks. 
However, conceptual knowledge, one essential kind of knowledge for human cognition, still remains understudied in this line of research. This limits PLMs' performance in scenarios requiring human-like cognition, such as understanding long-tail entities with concepts. 
In this paper, we propose ConcEPT, which stands for \textbf{Conc}ept-\textbf{E}nhanced \textbf{P}re-\textbf{T}raining for language models, to infuse conceptual knowledge into PLMs. 
ConcEPT exploits external taxonomies with entity concept prediction, a novel pre-training objective to predict the concepts of entities mentioned in the pre-training contexts.
Unlike previous concept-enhanced methods, ConcEPT can be readily adapted to various downstream applications without entity linking or concept mapping.
Results of extensive experiments show the effectiveness of ConcEPT in four tasks such as entity typing, 
which validates that our model gains improved conceptual knowledge
with concept-enhanced pre-training.

\end{abstract}

\section{Introduction}

% 预训练语言模型，language modeling预训练，无监督，大语料库
Pre-trained language models (PLMs) like BERT~\citep{devlin2018bert} and RoBERTa~\citep{liu2019roberta} have demonstrated impressive proficiency in language representation learning via unsupervised pre-training on massive corpora.
% PLMs很成功，很多NLP任务，证明他们学到了很多语言知识，
They have been prevailing in state-of-the-art methods for natural language processing (NLP) tasks% and become the new paradigm
, which reflects their superior ability in grasping linguistic knowledge.
% 还有世界知识 / 事实知识。
Furthermore, PLMs encode certain world knowledge in their parameters~\cite{petroni2019lama},
% 所以PLMs也被用在knowledge-intensive tasks上
and have been increasingly applied to more challenging knowledge-intensive tasks like entity typing~\citep{zhang2019ernie} and relation classification~\citep{soares2019matching}. 
However, there is still a gap between PLMs and human beings in terms of knowledge~\citep{kassner-schutze-2020-negated}.

% 进一步，knowledge-intensive tasks，KEPLM出现
To advance the performance of PLMs in knowledge-intensive tasks, knowledge-enhanced pre-trained language models (KEPLMs) are proposed, 
% 具体考虑了什么知识 
integrating various kinds of knowledge including entities~\citep{zhang2019ernie}, facts (relational triples) in knowledge graphs (KGs)~\citep{wang-etal-2021-kepler}, syntax~\citep{bai2021syntaxbert}, retrieved texts~\citep{guu2020realm} and logical rules~\citep{betz-etal-2021-critical}.  
% 然而 有一类重要知识被understudied了：概念知识
However, one essential kind of knowledge for human cognition - conceptual knowledge - still remains understudied in this line of research.

\begin{figure}[t]
    \centering
    \includegraphics[width=\columnwidth]{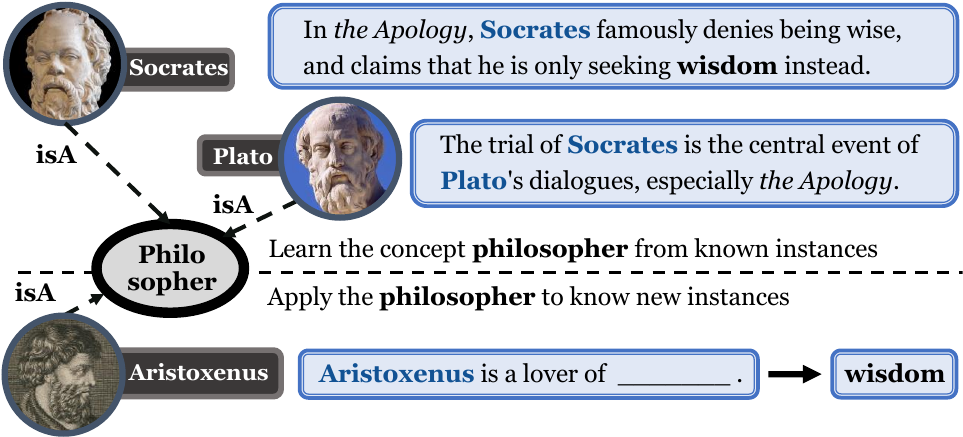}
    \caption{Examples of human-like learning and application of the concept \textit{philosopher}.}
    \label{fig:philosopher}
\end{figure}

% xyh0111
% done 这里的过度还要提升：concept对于知识密集型的任务是不可或缺的，
% done 还要呼应一下第二段
% 前后端的首句要能连在一起；第一句得是，概念对knowledge intesnive 任务很重要，或者和现有第一句连起来

% 没有概念知识，现存模型和人类有差距
Conceptual knowledge is indispensable to bridging the gap between existing PLMs and human-level  performance on many knowledge-intensive tasks.
%There are still gaps between existing PLMs and human-level  performance on many knowledge-intensive tasks, for which conceptual knowledge  is indispensable.
% 概念知识对人很重要
Concepts are the glue that holds our mental world together~\citep{murphy2004bigbook}.
%, and are thus indispensable for human-level performance on knowledge-intensive tasks.
% 举例：概念和实体间天然适合知识迁移
From \textit{Socrates} and \textit{Plato}, human learners learn the concept \textit{philosopher}, which facilitates their later understanding of other philosophers like \textit{Aristoxenus}, as illustrated in Figure ~\ref{fig:philosopher}.
% 概念和实体间天然适合知识迁移
In other words, knowledge of a concept and its instances can naturally be transferred to each other. 
% 我们对大量unpopular entity的理解，完全基于概念
Even though we know popular entities like \textit{Plato} well, there are abundant unpopular entities like \textit{Aristoxenus} that we know little about and understand mainly based on its concept \textit{philosopher}. 
% 这种基于概念的知识迁移，对PLM同样重要
Such concept-based knowledge transfer is just as important for PLMs, because these unpopular entities, or long-tail entities in other words, are challenging in many NLP tasks for data-driven models including PLMs~\cite{wang2022language}.
Clearly, conceptual knowledge is vital for human-like understanding of these entities.
% 以上指出了概念对PLM重要的一种原因，是否还有其他原因？暂时不管了

% taxonomic knowledge/concept property knowledge的区别，在intro里压缩一下，简单举例。
%第三段到绿色前就可以停了。概念知识的分类，往后放
%第四段，开头目前PLM没有概念知识。Concept对PLM很重要可以合进第三段。
% 第一句应该是PLMs没有概念知识，所以在什么任务上效果不好。把概念知识对PLM很重要倒过来。

% 0120: 立场不对，不能说LM已经学到了coarse concepts Done
% 现有研究只学到了coarse concepts ，基本概念， Done
% 但 but 后面更重要 对大量存在的细概念没学到 Done

% 然后 concept property 只学到了一些basic的，对抽象概念的，就很难学到 remove

% LM没学到什么概念？细概念 / 非基本概念 / 抽象概念；细概念的属性；没有学到概念层级关系； Considered

% 长尾实体和概念的isinstance；复杂概念的层级关系 Done

% 现有LM没有学到概念知识；虽然现有研究；但我们认为；总体而言，LM显著缺乏概念知识；长尾实体的概念；Done

% suffer from inaccuracy - 改成 现有的概念知识是从数据里学到的 data-driven的 分布式的隐性表示 由于数据有噪音 学习过程有错误 这个表示是有错误的 相对符号知识库 （放最后）

% 引用COPEN的发现：语言模型对概念认知不好
Existing PLMs systematically lack conceptual knowledge~\citep{peng2022copen}.
Although ~\citet{michael2020asking} suggests that language representations of PLMs could entail coarse concepts like \textit{person} and \textit{date}, their latent conceptual knowledge is often inaccurate compared with human-curated taxonomies~\citep{aspillaga-etal-2021-inspecting}, and they significantly  struggle with fine-grained concepts and concept hierarchy~\citep{peng2022copen}, let alone the ability to conceptualize long-tail entities.
%existing PLMs still suffer from inaccuracies of their latent conceptual knowledge~\citep{aspillaga-etal-2021-inspecting}, 
Without knowledge of the underlying concepts, \textit{Plato} and \textit{Aristoxenus} would be viewed as irrelevant tokens, and PLMs make no knowledge transfer in between.

%First, knowledge of the underlying concepts plays a vital role in knowledge-intensive tasks like relation classification~\citep{peng-etal-2020-learning}, commonsense reasoning~\citep{liu2021kgbart} and question answering~\citep{garderes-etal-2020-conceptbert}.

% 0120 加一个帽子：没有人将概念引入PLM。然后再现有工作主要是下游任务
% infuse conceptual knowledge into PLMs at the pre-training stage,

% Hence, 外部概念知识也被多种NLP / Knowledge Intenstive 任务引入
It remains understudied to infuse conceptual knowledge into PLMs. 
External conceptual knowledge, including concept-oriented KGs like ConceptNet~\citep{speer2017conceptnet} and taxonomies like Probase~\citep{wu2012probase}, has been introduced to boost PLMs' performance in downstream tasks including relation classification~\citep{peng-etal-2020-learning}, commonsense reasoning~\citep{liu2021kgbart} and question answering~\citep{garderes-etal-2020-conceptbert}. 
These works have demonstrated the importance and effectiveness of conceptual knowledge in downstream applications. 
% 这么多任务都需要概念知识，那就应该在在预训练端注入概念知识
Considering the universal importance of conceptual knowledge, 
it is clearly beneficial to
infuse conceptual knowledge into PLMs at the pre-training stage, which can fundamentally advance PLMs' knowledge, ease downstream efforts, and support a broader range of applications.    

% 0118 现在没有工作往预训练模型里注入概念，we argue that ，会有什么好处 Done
% 0118 interested这个词改掉 Done

%As an example, visual concepts derived from object detection have also been leveraged to pre-train vision-language models~\citep{tan-bansal-2019-lxmert, li2022mvptr}, and proved to be effective.
%However, conceptual knowledge for PLMs remains understudied.
%It remains understudied to enrich language model pre-training with conceptual knowledge. 

%\polish{
%There are two kinds of concept-related knowledge:  
%In this paper, we roughly divide conceptual knowledge into two classes, including  
%discuss two types of conceptual knowledge, 
%taxonomic knowledge such as \textit{Plato is a philosopher}, and concept property knowledge such as \textit{Birds can fly}. 
%}
%\polish{
% 概念层级知识
%Taxonomic knowledge (isA relations) 
%captures hierarchical relationships among concepts and their instances, 
%tells whether an entity is an instance of a concept or a concept is a subclass of another
%such as \textit{Plato is a philosopher}. 
% 概念属性知识
%Concept property knowledge tells typical properties of concepts, which often apply to their typical instances, such as \textit{Birds can fly}.}

% 我们提出了ConcEPT，概念增强的预训练模型，（手段）利用taxonomy（目标）提高PLM的概念理解
In this paper, we propose \textbf{Conc}ept \textbf{E}nhanced \textbf{P}re-\textbf{T}raining for language models (ConcEPT), which exploits external taxonomies to enhance PLMs with conceptual knowledge. 
% ConcEPT具体怎么做：概念预测，使用isA 关系 作为概念知识
ConcEPT introduces a novel pre-training objective named entity concept prediction (ECP), %which leverages concepts for self-supervised pre-training, 
predicting the concepts of entity mentions based on their contexts.  
% ConcEPT基于taxonomic knowledge，从taxonomy来
This objective is supported by taxonomic knowledge, namely isA relations from external taxonomies. 
% 让概念成为连接corpus的桥梁，就像概念连接人类的心理世界一样
% xyh 扮演了一种桥梁的作用，建立起了2个不相关的token间的语义关联（这个解释就不民科了）。Done
% 然后要举个例子，比如“苏格拉底”影响了“哲学家”表示，哲学家又约束了“柏拉图”表示。 Done
With ECP, language representations of different entities with a common concept, e.g., \textit{Plato} and \textit{Aristoxenus}, are semantically correlated with a shared objective.
For downstream applications, the pre-trained ConcEPT model can be readily applied as BERT with no additional model modification or data pre-processing.
We evaluate ConcEPT with four typical knowledge-intensive tasks, including entity typing, conceptual knowledge probing, relation classification and knowledge graph completion. 
Their results show that ConcEPT acquires improved conceptual knowledge, including taxonomic knowledge, concept property knowledge and the capability of concept-based knowledge transfer.
%, which is just like that concepts serve as the glue that holds our mental world together. 这句太民科
% 是否是想证明，概念增强使得泛化性更好？
%This correlation helps models to better generalize from learned samples to unseen ones if there are underlying concepts in common.}
%For example, ECP learns the concept label \textit{philosopher} from \textit{Plato}, and afterwards encourages the representation of \textit{Aristoxenus} to match the concept label \textit{philosopher}. 放到method讲吧
%\polish{Thereafter, isA relations serve as the link that connects unstructured pre-training corpus, 
%just as that concepts serve as the glue that holds our mental world together.}
% 再具体一点：预训练时需要实体链接、背景知识图谱，下游任务中不需要
In summary, our contributions are listed as follows:
\begin{enumerate}
    \item %For the general importance of conceptual knowledge in  knowledge-intensive tasks, we propose concept-enhanced pre-training. 
    We propose concept-enhanced pre-training, with a novel pre-training objective, entity concept prediction, to exploit external taxonomies and use concepts as supervision. 
    To the best of our knowledge, we are the first to infuse conceptual knowledge into PLMs at the pre-training stage.  
    \item We conduct extensive experiments to evaluate our ConcEPT model. 
    The results demonstrate that ConcEPT outperforms existing KEPLMs in entity typing, and achieves improved performance on three other knowledge-intensive tasks, which validates that ConcEPT acquires enhanced conceptual knowledge. % with concept-enhanced pre-training. 
    
\end{enumerate}

%We conduct extensive experiments to validate ConcEPT's acquisition of both kinds of concept-related knowledge, and show its improvement on various other knowledge-intensive tasks.
%Our extensive experiments validate ConcEPT's acquisition of both taxonomic knowledge (like \textit{Plato is a philosopher}) and concept property knowledge (like \textit{Birds can fly}), and show its improvement on various other knowledge-intensive tasks.
% 首先，Concept-related tasks，超越各种KEPLM；Knowledge-intensive tasks，也带来了提升
%That is, ConcEPT not only outperforms existing KEPLMs on concept-oriented tasks like entity typing and concept property judgment, but also bring significant improvement over a wide range of other knowledge-intensive tasks, including relation classification, triple classification, and link prediction.

\section{Related Works}

\subsection{Knowledge Enhanced Pre-trained Language Models}
% 这一段2句话 1句讲BERT T5。第二句话讲他们有什么问题

% 介绍PLM
Pre-trained language models like BERT~\citep{devlin2018bert} and T5~\citep{raffel2020t5}  have established new state-of-the-art in many NLP tasks. % for their powerful language representation. 
% knowledge-intensive tasks 
However, as their language modeling objectives do not direct toward world knowledge, they still struggle with knowledge-intensive tasks. %~\citep{wei2021knowledge}. 
% 介绍KEPLMs
Hence, many efforts are made to inject world knowledge into PLMs%towards better capacity in knowledge-intensive tasks
, which are referred to as KEPLMs.

% 我们主要集中于用KG增强的
A major line of study in KEPLMs is to integrate knowledge from KGs, mainly by knowledge interpolation and/or knowledge supervision.
%\polish{Specifically, we mainly focus on KEPLMs that incorporates fact knowledge in KGs.} 
Knowledge interpolation is to supplement input sentences by inserting related knowledge like entities~\citep{zhang2019ernie} or relational triples~\citep{liu2020kbert, sun2020colake}.
%such as ERNIE~\citep{zhang2019ernie}, K-BERT~\citep{liu2020kbert} and CoLAKE~\citep{sun2020colake}, directly augment the input sentences by inserting related  entities or relational triples.  
%ERNIE~\citep{} integrates fact knowledge in KGs into PLMs, which are encoded into entity embeddings via 
%knowledge representation learning methods like TransE~\citep{}. It also proposes dEA, a pre-training object that requires PLMs to predict masked token-entity alignments. 
%K-BERT~\citep{} and CoLAKE~\citep{} directly insert relational triples into related sentences.
%They modify position embeddings and visible attention matrix
%However, expansion / select triples .
This requires entity linking on not only the pre-training corpus, but also downstream datasets, which makes them inconvenient to use.
Knowledge supervision designs knowledge-guided pre-training objectives, such as knowledge-guided masked language modeling~\citep{erniebaidu, shen-etal-2020-exploiting}, entity linking~\citep{zhang2019ernie, peters-etal-2019-knowledge}, knowledge graph completion~\citep{wang-etal-2021-kepler, qin-etal-2021-erica} and text generation from KG subgraphs~\citep{agarwal-etal-2021-knowledge}.
%guide mask (ERNIE Tsinghua), 
%predict entity , KnowBERT/ERNIE (
%triple classification, KEPLER,
Besides, there are also KEPLMs enhanced by other types of knowledge, such as syntax trees~\citep{bai2021syntaxbert},  retrieved texts~\citep{guu2020realm} and logical rules~\citep{betz-etal-2021-critical}.

\subsection{Conceptual Knowledge and Pre-trained Models}
\paragraph{\textbf{Conceptual Knowledge in PLMs.}}
Previous works %have studied conceptual knowledge in PLMs. Their findings 
suggest that PLMs' language representations group together based on semantics~\citep{NEURIPS2019_159c1ffe} and entail coarse concepts like \textit{person} and \textit{date}~\citep{michael2020asking}.
%这局下面有了
%However, existing PLMs are still indifferent to fine-grained concepts~\citep{michael2020asking}.
~\citet{dalvi2022discovering, aspillaga-etal-2021-inspecting} extract latent concepts and taxonomies from PLMs, which still pale in comparison with human-curated knowledge in terms of accuracy and coverage. 
~\citet{wang2022prototypical} shows that PLMs can well judge taxonomic relations after fine-tuning.
~\citet{petroni2019lama} demonstrates that PLMs acquire basic concept property knowledge like \textit{Birds can fly}.
%, and even entail hierarchical relationships among them~\citep{aspillaga-etal-2021-inspecting}.
%~\citet{dalvi2022discovering} groups the representations into hierarchical clusters as latent concepts of PLMs. 
%~\citet{NEURIPS2019_159c1ffe} shows that language representations from PLMs group together in the representation space based on semantic meanings. 
%~\citet{michael2020asking} further demonstrates the representations distinguish coarse concepts like \textit{person} and \textit{date}, but are indifferent to fine-grained concepts. 
%~\citet{aspillaga-etal-2021-inspecting} probes whether the representations encode hierarchical relationships among concepts, and extracts a latent taxonomy from the representations. 
~\citet{peng2022copen} constructs a benchmark to measure conceptual knowledge in PLMs, and shows their inability to distinguish fine-grained concepts and understand concept hierarchy. 
% based on taxnomies
%It contains three tasks, which respectively 
%questioning PLMs' capacity to recognize conceptually similar entities, learn concept properties, and distinguish the most appropriate concepts in contexts. 
% 备选：challenging both taxonomic knowledge and concept property knowledge of PLMs

\paragraph{\textbf{Conceptual Knowledge for PLMs.}}
Conceptual knowledge, including taxonomic knowledge and concept property knowledge, has been widely applied to enhance PLMs in downstream tasks.
% Taxonomic
Taxonomic knowledge is represented as \textit{isA} relations in KGs, especially taxonomies like Probase~\citep{wu2012probase}. 
It identifies the underlying concepts behind input sentences and entities, thus improving PLMs % to better understand entities and make decisions 
in tasks like relation classification ~\citep{peng-etal-2020-learning} and KG alignment~\citep{zeng2021encoding}.    
%~\citet{peng-etal-2020-learning} shows that PLMs largely rely on concept information of entities in relation classification tasks. 
%Besides, taxonomic knowledge are also applied to KG alignment ~\citep{zeng2021encoding}, visual question answering  
%like relation classification~\citep{peng2020learning}, commonsense reasoning~\citep{liu2021kgbart} and . 
% 概念关系知识 理解和推理概念
Concept property knowledge is provided by concept-oriented KGs like ConceptNet~\citep{speer2017conceptnet}.
It helps language models to better understand and reason over concepts in tasks like commonsense generation~\citep{liu2021kgbart, bosselut-etal-2019-comet} and question answering ~\citep{wang-etal-2020-connecting, garderes-etal-2020-conceptbert}. 

%Concept-oriented knowledge graphs like ConceptNet~\citep{speer2017conceptnet} contain plentiful conceptual knowledge, 
%KG-BART~\citep{liu2021kgbart} reasons over concept property knowledge in ConceptNet~\citep{speer2017conceptnet} to generate commonsensible sentences from given concepts.
%COMET~\citep{} trains a Transformer model with triples in ConceptNet to generate commonsense concept knowledge in natural language. 

Conceptual knowledge has also been considered to pre-train better PLMs, but none of these efforts targets improved concept understanding of PLMs. 
WKLM~\citep{Xiong2020Pretrained} learns entities by judging whether they are replaced with conceptually similar ones that contradict the contexts. 
~\citet{lee2022ccm} proposes concept-based curriculum masking to accelerate pre-training.
%proposes concept-based curriculum masking to accelerate pre-training. %, which gradually masks concept words related to previously masked ones. 
They neglect to use concepts as supervision. 
Researches on vision-language models have introduced concept-supervised objectives for pre-training~\citep{tan-bansal-2019-lxmert, li2022mvptr}.
%Concepts have been introduced to serve as supervision in pre-training vision-language models~\citep{tan-bansal-2019-lxmert}~\citep{li2022mvptr}. 
However, their concepts are derived from image features via object detection, instead of human-curated knowledge.
In this paper, we enhance PLMs' conceptual knowledge with supervision from real-world taxonomies.

%They use object detection methods like Faster R-CNN~\citep{ren2015fasterrcnn} to obtain concepts of image regions, and pre-train their models to predict these concepts given these regions. 
% 概念增强之于文本模型和之于视觉模型是不同的
%Although they also learn concept prediction as a pre-training objective, their concept prediction in images is totally different from our concept prediction in texts. 
%The reasons are mainly twofold.
%First, different image regions of a concept's entities are visually similar, while text mentions of a concept's different entities are mostly dependent. Therefore, concept prediction in images expects models to learn the concept's own visual features, while concept prediction in texts requires models to not only remember names of the concept, but also learn the concept's textual contexts. 
%Second, they obtain concepts of image regions by object detection, while we retrieve concepts of entity mentions by knowledge graphs.

\section{Concept-Enhanced Pre-training}
% ConcEPT是什么
In this section, we elaborate ConcEPT, which exploits external taxonomies to enhance its conceptual knowledge. 
% 这一章 我们介绍ConcEPT的framework和details
%In this section, we elaborate concept-enhanced pre-training objective and detailed implementation of ConcEPT. 
We first introduce entity concept prediction, a concept-supervised objective for pre-training in Sec 3.1. 
Then, we describe our concept selection and taxonomy construction in Sec 3.2, and other implementation details in Sec 3.3. 

% 0120 需要把框架图先讲一下，entity linking，讲一下整个famework Done

\subsection{Entity Concept Prediction}

% Motivation：让概念和实体interlinked；这个假设是不是太强？先按自己的理解说吧
For human cognition, knowledge about a concept and its instances are tied together~\citep{murphy2004bigbook}. For example, we learn the concept \textit{philosopher} from its typical instances like \textit{Socrates}, 
and apply this concept to understand its others instances like \textit{Aristoxenus}.
%also know about unfamiliar ones like \textit{Aristoxenus} based on the concept \textit{philosopher}. 
% 现存的语言模型无法捕捉这样的关联，因为没有underlying concepts
Existing PLMs hardly capture such interaction, without knowledge of the underlying concepts. 
% 所以用外部概念知识
Hence, we leverage external taxonomies to enhance conceptual knowledge of PLMs.
%taxonomic knowledge, namely \textit{isA} relations among entities and concepts.

\begin{figure*}[htb]
    \centering
    \includegraphics[width=\linewidth]{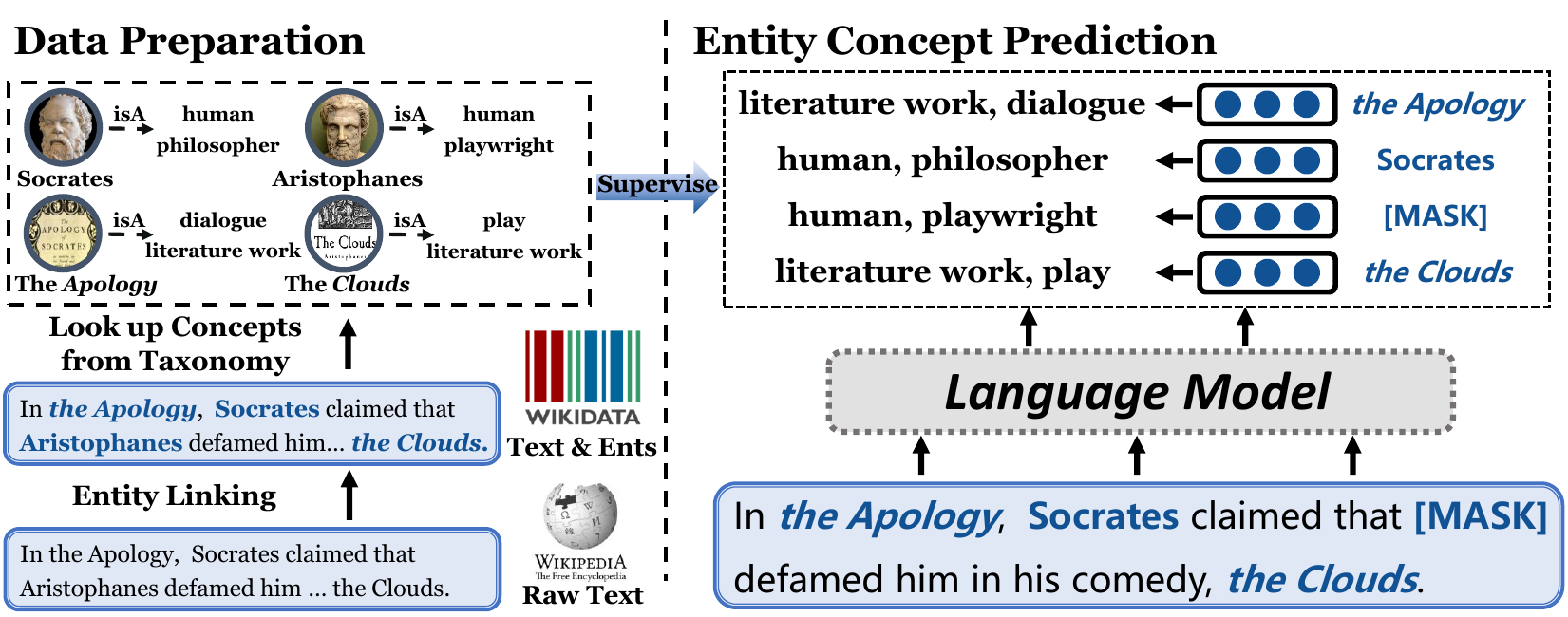}
    \caption{The framework of pre-training with entity concept prediction. The annotated entity mentions are in blue. }
    \label{fig:ecp}
\end{figure*}

% 概括：我们的工作 / ConcEPT 提出了一个全新的预训练目标：实体概念预测。
We propose \textbf{entity concept prediction (ECP)}, a novel pre-training objective to infuse conceptual knowledge into PLMs.
Its framework  is illustrated in Figure ~\ref{fig:ecp}. 
We first annotate entities in input documents via entity linking, then look up their concepts from taxonomies, and require PLMs to predict these concepts.
% 实体概念预测要求预测实体对应的概念
%which requires PLMs to predict the concepts of entities mentioned in input documents.
% 概念从何而来：外部知识库 taxonomy
%The concepts are provided by external taxonomies as knowledge source.
% 具体来说，ECP怎么做
Specifically, 
% 建模：
% 给定输入文本
given an input sequence $(x_{1}, \ldots, x_{n})$ of $n$ tokens, 
% 和标注出的实体mention
and an entity mention $m=(e, i, j)$ indicating the span $(x_{i}, \ldots, x_{j})$ is linked to entity $e$,
% 概念预测是
ECP aims to predict $e$'s concepts $\mathcal{C}_{e}\subseteq\mathcal{C}$, where $\mathcal{C}$ denotes the candidate concepts.
%, judging whether each $c\in\mathcal{C}$ is a concept of $e$, where $\mathcal{C}$ is the candidate concept set.
% 实体可能被mask
Entity mentions are masked at random, and PLMs may need to predict concepts based solely on the contexts, without entity names.  
% 技术上，实体如何表示：boundary
We represent the entity mention $m$ with final representations $ \textbf{h}_{i}, \textbf{h}_{j}\in\mathcal{R}^d$ of its boundary tokens $x_{i}$, $x_{j}$ from PLMs, where %$l$ denotes the number of layers of PLMs and 
$d$ is the dimension. 
%$\textbf{e} = [\textbf{x}_{i}, \textbf{x}_{j}]$, namely the concatenation of the final representations of the boundary tokens $\{x_{i}, x_{j}\}$ from PLMs.
% 然后，如何预测概念？分类
Afterward, we train ConcEPT to predict $\mathcal{C}_{e}$ via binary classification whether each $c\in\mathcal{C}$ is a concept of $e$.

% 对于ECP 需要引入额外参数 一个ECP Layer
We add an ECP head on top of entity representations for ECP learning.  
%which includes a conceptualization layer and a merging layer.
This head is a two-layer MLP, similar to the MLM head used in BERT: 
\begin{equation}
    \hat{\textbf{y}}_m = \sigma(\textbf{W}_c(\phi(\textbf{W}_u[\textbf{h}_i;\textbf{h}_j] + \textbf{b}_u)) + \textbf{b}_c),
\end{equation}
where $\hat{\textbf{y}}_m\in\mathcal{R}^{|\mathcal{C}|}$ is the ECP output, $[\cdot;\cdot]$ denotes concatenation, $\phi(\cdot)$ and $\sigma(\cdot)$ are GELU activation and sigmoid function, and $\textbf{W}_u$, $\textbf{b}_u$, $\textbf{W}_c$, $\textbf{b}_c$ are learnable weights and biases.
%\polish{The $k$-th row in  $\textbf{W}_c\in\mathcal{R}^{|\mathcal{C}|\times 2d}$, viewed as representation of the $k$-th concept $c_{k}$, is initialized by word embeddings of its name.}
The $k$-th element of $\hat{\textbf{y}}_m$ is the probability $P(c_k|m)$ that the $k$-th concept $c_k$ is a concept of $m$.
Finally, this objective is optimized with binary cross-entropy loss: 
%\begin{equation}
%    \mathcal{L}_{\text{ECP}} = \sum_{c\in\mathcal{C}} (\mathbbm{1}_{c\in\mathcal{C}_e}\text{log}P(c|m) + \mathbbm{1}_{c\notin\mathcal{C}_e}\text{log}(1-P(c|m)) )
%\end{equation}
\begin{equation}
    \mathcal{L}_{\text{ECP}} = \sum_{c\in\mathcal{C}_e} \text{log}P(c|m) + \sum_{c\notin\mathcal{C}_e} \text{log}(1-P(c|m)) 
\end{equation}
% 0120 ：写事实性的东西。实验证实了，有ECP，可以做conceptual based knowledge tr
% 0120 这个太虚了 不如删掉 ... 肖老师说可以留
% 0120 需要有一段篇幅去解释Figure 2；或者放Figure 2标题里 Done

% 有了ECP，概念和实体就被联系起来了，因为实体表示需要能持续地match概念表示，
%\polish{With ECP, concepts and their instances are interlinked, because representations of entity mentions from PLMs should consistently match with corresponding concept representations so as to accomplish ECP.
% 另一方面，从实体的上下文能预测出概念，也是非常合理的

% 0119 不如变成公式2为什么合理的解释。P(c|m)
% Y-M P(c|m) 
% 最简单的绿色 没有意义，可以都不要。有点画蛇添足
% 最好放的 其实是公式的合理性，补充的东西。
% 因为m可以同时属于多个概念，选择BCE而非softmax。
% 为什么概念预测是合理的：作为一个自监督的任务。（监督大量，相对高质量 - 没必要）
% 还是删光吧。

%\polish{
%Since concepts are generalization of entities, it is also highly reasonable that a concept is related to and inferable from many of its entities' contexts.

%For a concept $c$ and its instance $e$, since $c$ is a superordinate of $e$, it is highly  reasonable that the concept $c$ is related to and can be inferred from many of $e$'s contexts. 

% 概念提供了更多的监督信号
%Concepts provides plentiful reasonable supervision signals, regularize representation space.
%}

% Jiaqing : advantage 。概念从哲学上是有意义的，实践也是有意义的。训练方式是可行的 Done。至于怎么工作，就不管了，不是能证明的。

By using concepts as supervision, we integrate external taxonomies into PLMs. Hence, ConcEPT learns concept-aware language representations,  especially that of entities which entail concept information. Additionally, this enables different entities of one concept, such as \textit{Plato} and \textit{Aristoxenus}, to be semantically correlated in the representation space, which simulates human cognition.

%representations of entities encode the concept information, and different entities of one concept,  e.g., \textit{Plato} and \textit{Aristoxenus}, are semantically correlated in the representation space. %which facilitates concept-based knowledge transfer. 
%We show with experiments in Sec \ref{sec:exp} that ConcEPT acquires advanced conceptual knowledge, including taxonomic knowledge, concept property knowledge, and the capacity of this concept-based knowledge transfer.

% 0120 上面这一段 intro出现过了 不需要了

\subsection{Concept Selection and Taxonomy Construction}
% 先说Concept选择，BLC。这一段无关具体技术细节。（corpus+kg）
% 我们选择热门的、basic level的概念
We select popular and basic-level concepts for ECP learning.
% 一方面，为什么要popular
On one hand, popular concepts are frequent in texts and thus provide plentiful supervision. 
% 另一方面，为什么选BLC：比起fine-grained concept，更与上下文相关，适合预测，provides better supervision
On the other hand, people tend to categorize entities with basic-level concepts
%basic-level concepts are concepts which people most likely categorize entities as
~\citep{murphy2004bigbook}. For PLMs, basic-level concepts are deeply associated with contexts of their entities, so they serve as better supervision compared with more complex concepts.  
For example, the context "\textit{Plato} argues that knowledge is distinguished from mere true belief" is associated with \textit{Plato}'s basic-level concept \textit{philosopher}, but not necessarily its complex concept \textit{political philosopher}. % which would be noise 
% basic和popular是高度相关的
The popularity and basicness of concepts are correlated, since basic-level concepts tend to be popular and vice versa.

% xyh0111
% 哲学家不是一个很好的基本概念的例子，用猫或狗更好

\begin{table}[h!]
    \begin{center}
    \centering
    \setlength{\tabcolsep}{0.4mm}{
    \resizebox{\columnwidth}{!}{
    \begin{tabular}{ccccc}
    \Xhline{0.8pt}
        \textbf{Property} & & \textbf{Alias} & & \textbf{Example}  \\
    \Xhline{0.8pt}
        P31 & & \textit{instance of} & & (\textit{Nelson Mandela}, P31, \textit{human}) \\
        P279 & & \textit{subclass of} & & (\textit{apple}, P279, \textit{fruit}) \\ 
        P136 & & \textit{genre} & & (\textit{The Republic}, P136, \textit{dialogue}) \\ 
        P106 & & \textit{occupation} & & (\textit{Plato}, P106, \textit{philosopher}) \\ 
        P7736 & \ & \textit{form of creative work} & \ \ \  & (\textit{Abbey Road}, P7736, \textit{studio album})
         \\ 
    \Xhline{0.8pt}
    \end{tabular}}}
    \end{center}
        \caption{Properties considered as \textit{isA} relations. }
    \label{tab:isA_relations}
\end{table} 

% 先总结，我们构造了一个基于wikidata的taxonomy用以预训练
In this paper, we construct a Wikidata-based taxonomy, WikiTaxo, to support concept-enhanced pre-training. 
A taxonomy is composed of \textit{isA} triples ($e$, \textit{isA}, $c$) stating that entity $e\in\mathcal{E}$ belongs to concept $c\in\mathcal{C}$, where $\mathcal{E}, \mathcal{C}$ are the entity set and the concept set. 
% 由于变成taxonomy构造了，先讲entity set 最简单的
We first obtain $\mathcal{E}$ by 
annotating entities mentioned in Wikipedia, since Wikipedia is a popular pre-training corpus, contains lots of real-world knowledge, and deeply matches Wikidata. The entities are annotated via Tagme~\citep{ferragina2010tagme} entity linking.
% 具体来说，我们是选择了wikidata中的5种relation
Then, we consider five types of \textit{isA} relations in Wikidata  listed in Table \ref{tab:isA_relations}\footnote{The relations are selected according to \url{https://www.wikidata.org/wiki/Property\_talk:P31}, where replacement properties of P31 (\textit{instance of}) are enumerated. P136 triples with human subjects are excluded.}, 
and collect \textit{isA} triples of entities in $\mathcal{E}$.

% 然后，选择概念，popular lexical，这里的single token就比较隐晦
Afterward, we obtain $\mathcal{C}$ by  selecting popular lexical concepts from candidate concepts in the \textit{isA} triples.  
We measure a concept's popularity with two metrics: 
(1) Searching frequency of the concept (SFC). A popular concept is supposed to be frequently searched by Internet users. We obtain SFC from Wikidata QRank~\footnote{\url{https://github.com/brawer/wikidata-qrank}}, which aggregates page view statistics for Wikipedia and other Wikimedia projects.   
(2) Mentioning frequency of its entities (MFE). A popular concept tends to have well-known and frequently mentioned entities. We first count frequency of each entity mentioned in Wikipedia,
%traverse the corpus to count entity mentions, 
and then calculate a concept's MFE by summing over its entities.  
We keep lexical concepts with MFE above 100,000 or MFE above 10,000 and SFC above 1,000,000. 
% 建立BLC和single token的联系
We further select lexical concepts that can be referred to by individual words, which are closely associated with basic-level concepts~\citep{sepconcept} %that can be referred to by individual words in natural languageconcepts that can be tokenized into single words 
~\footnote{In this paper, we use BERT tokenizer.
%BERT as the base PLM and its tokenizer. For PLMs with fine-grained tokenizers, lexical concepts can be distinguished via splitting around space. 
}. 
Furthermore, we invite annotators to manually process the concepts by removing noisy and non-informative ones while merging similar ones. 
Eventually, we obtain 1305 popular and basic-level concepts.

% 0120 还是给一个准则和依据；
% 只有1个人，当吃不准的时候，再请一个人。
% annotator，丢在附录地方

% 先声明，我们基于wikipedia+wikidata
%We pre-train ConcEPT on Wikipedia corpus~\footnote{\url{https://en.wikipedia.org/}}, linked  with Wikidata~\footnote{\url{https://wikidata.org/}} entities. 
% 因此，我们的taxonomy构建和概念选择，亦是基于wikidata
%Consequently, the collection of our concepts and taxonomy are based on Wikidata. 

\subsection{Implementation Details}

\paragraph{\textbf{Pre-training Data.}}
% Corpus: Wikipedia
%ConcEPT uses English Wikipedia as the pre-training corpus.
% 先声明，我们基于wikipedia+wikidata
We pre-train ConcEPT on the Wikipedia corpus~\footnote{\url{https://en.wikipedia.org/}}, linked  with Wikidata~\footnote{\url{https://wikidata.org/}} entities. 
% dump, WikiExtractor
We obtain Wikipedia texts from the April 2021 data dump, extracted via WikiExtractor\footnote{\url{https://github.com/attardi/wikiextractor}}. 
% Wikidata Tagme
The entities in texts are annotated via Tagme, with a score threshold of 0.3. %~\cite{ferragina2010tagme}.
% Taxonomy, relations, simpple-wikidata-db
We extract Wikidata from its dump of 20221007 via simple-wikidata-db\footnote{\url{https://github.com/neelguha/simple-wikidata-db}}.%, from which 

\paragraph{\textbf{Model Architecture.}}  
% 结构：Transformer Encoder
We implement ConcEPT based on the Transformer~\citep{vaswani2017attention} encoder architecture. 
% 具体：BERT一样的结构
We adopt the same architecture as BERT-base, including 12 Transformer blocks and a two-layer MLM head. 
% 参数初始化
We initialize ConcEPT with pre-trained parameters of BERT, 
% ECP 
and add the ECP head.
Our implementation is based on HuggingFace  Transformers~\citep{wolf2019huggingface}.

\paragraph{\textbf{Pre-training Objective.}} 
We pre-train ConcEPT with the ECP objective and the masked language modeling (MLM) objective.
% 对所有实体进行概念预测
For concept-enhanced learning, we predict the concepts of every entity mention in the texts, 
% 15%实体被遮盖
while 15\% of entity mentions are masked entirely. %to force ConcEPT to predict the concepts based solely on the contexts. 
For settings of MLM, we follow BERT~\cite{devlin2018bert}. 

\paragraph{\textbf{Hyperparameters.}} 
We optimize ConcEPT with AdamW, with learning rate 2$e$-5 and weight decay 1$e$-2.
We set the batch size to be 256, and the maximal input length to be 512. 
We pre-train ConcEPT on a single 32G NVIDIA GeForce RTX 3090 for 100,000 steps, which takes around 240 hours. 
Other parameters follow BERT~\cite{devlin2018bert}.

\section{Experiments}
\label{sec:exp}
In this section, we evaluate ConcEPT's performance on knowledge-intensive tasks, and analyze its conceptual knowledge.
Our experiments cover four kinds of tasks including entity typing, conceptual knowledge probing, relation classification and knowledge graph completion. They concern three aspects of conceptual knowledge: taxonomic knowledge, concept property knowledge, and the capability of concept-based knowledge transfer. Statistics of our datasets are shown in Table \ref{tab:statistics}. 

%We are interested in ConcEPT's performance \polish{on four kinds of tasks, including tasks about conceptualization knowledge (Section ), tasks about concept property knowledge (Section ), tasks about ..., and other knowledge-intensive tasks}. 

\begin{table}[h]
    \centering
    \small
    \setlength{\tabcolsep}{0.3mm}{
    %\resizebox{\columnwidth}{!}{
    \begin{tabular}{*{6}{c}}
    \toprule
    \multirow{2}*{Dataset} & \multicolumn{3}{c}{Statistics} & \multicolumn{2}{c}{\multirow{2}*{\vspace{-0.2cm}\shortstack{Task-Specific\\ Information}}} \\
    \cmidrule{2-4}&  \#Train & \#Dev & \#Test & \multicolumn{2}{c}{} \\
        \midrule
    \multicolumn{4}{c}{Entity Typing} & \multicolumn{2}{c}{\#Type} \\
    \cmidrule(lr){1-4} \cmidrule(lr){5-6}
     FIGER & 2,000,000 & 10,000 & 563 & \multicolumn{2}{c}{113} \\
     FIGER-finer & 192,939 & 484 & 484 & \multicolumn{2}{c}{59} \\
     Open Entity & 1,998 & 1,998 & 1,998 & \multicolumn{2}{c}{9}\\
     \midrule
     \multicolumn{4}{c}{Conceptual Knowledge Probing} & \multicolumn{2}{c}{Form} \\
     \cmidrule(lr){1-4} \cmidrule(lr){5-6}
     CSJ COPEN & 4,462 & 1,116 & 3,909 & \multicolumn{2}{c}{Multi-Choice} \\
     CPJ COPEN & 3,274 & 823 & 4,758 & \multicolumn{2}{c}{Binary CLS} \\
     CiC COPEN & 2,888 & 722 & 2,368 & \multicolumn{2}{c}{Multi-Choice}\\
     \midrule
     \multicolumn{4}{c}{Knowledge Graph Completion} & \#Ent & \#Rel \\
     \cmidrule(lr){1-4} \cmidrule(lr){5-6}
     Wiki-CKT & 2,000 & 2,000 & 2,000 & 7,621 & 129  \\
     FB15k-237 & 272,115 & 17,535 & 20,466 & 14,541 & 237 \\
     \midrule
    \multicolumn{4}{c}{Relation Classification} & \multicolumn{2}{c}{\#Rel}\\
     \cmidrule(lr){1-4} \cmidrule(lr){5-6}
     %FewRel & 8,000 & 16,000 & 16,000 & \multicolumn{2}{c}{80}\\
     TACRED & 68,124 & 22,631 & 15,509 & \multicolumn{2}{c}{42}\\
    \bottomrule
    \end{tabular}}%}
    \caption{Statistics of the datasets used in this paper. 
    CLS stands for classification. 
    %CLS, ET, and RC denotes classification, entity typing, and relation classification respectively.
    }
    \label{tab:statistics}
\end{table}

\subsection{Entity Typing}

\begin{table*}[htbp]
    %\small
    \centering
    \resizebox{\textwidth}{!}{
    \begin{tabular}{cccccccccccc}
    \Xhline{0.8pt}
        \multirow{2}*{Dataset} &
        \multicolumn{4}{c}{\textbf{FIGER}} & 
        \multicolumn{4}{c}{\textbf{FIGER-finer}} & 
        \multicolumn{3}{c}{\textbf{Open Entity}}  
        \\
    \cmidrule(lr){2-5} \cmidrule(lr){6-9} \cmidrule(lr){10-12}
         & 
        \textbf{Acc} & \textbf{Ma-F} & \textbf{Mi-F} & \textbf{TMa-F} &
        \textbf{Acc} & \textbf{Ma-F} & \textbf{Mi-F} & \textbf{TMa-F} &
        \textbf{P} & \textbf{R} & \textbf{Mi-F} \\
    \Xhline{0.8pt}
        %\multicolumn{13}{|c|}{\textbf{Structure-based Knowledge Embeddings}} \\ \Xhline{0.8pt}
        %$\text{TransE~\cite{bordes2013transe}}{\clubsuit}$ & 0 & 0 & 0 
        BERT & 52.0 & 75.2 & 71.6 & 53.3 %& 52.04 & 75.16 & 71.63 
        & 92.6 & 94.4 & 93.6 & 91.9
        & 76.4 & 71.0 & 73.6 %& 76.37 & 70.96 & 73.56
        \\
        ERNIE & 57.2 & 76.5 & 73.4 & -   %& 57.19 & 76.51 & 73.39
        & - & - & - & -
        & 78.4 & 72.9 & 75.6 %& 78.42 & 72.90 & 75.56
        \\
        %RoBERTa${\spadesuit}$ & - & - & - & -
        %& - & - & - & -
        %& 77.4 & 73.6 & 75.4 
        %\\ \hline
        KEPLER & - & - & - & - 
        & - & - & - & -
        & 77.8 & 74.6 & 76.2
        \\
        KnowBERT & - & - & - & - 
        & - & - & - & -
        & \textbf{78.6} & 73.7 & 76.1
        \\
        CoLAKE & - & - & - & - 
        & - & - & - & -
        & 77.0 & 75.7 & 76.4
        \\
        BERT-contd & 57.5 & 76.3 & 73.5 &  55.7
        & 93.2 & 94.8 & 94.1 & 91.8
        & 77.8 & 73.4 & 75.5
        %& 0 & 0 & 0
        \\
        ConcEPT & \textbf{57.7} & \textbf{77.0} & \textbf{74.1} & \textbf{58.0}  %55.60 & 80.18 & 74.27
        & \textbf{94.2} & \textbf{95.4} & \textbf{94.7} & \textbf{93.2}
        & 76.3 & \textbf{76.6} & \textbf{76.4} 
        \\
    \Xhline{0.8pt}
    \end{tabular}}
    \caption{Results on FIGER, FIGER-finer and Open Entity(\%). 
    Ma-F and Mi-F denote loose macro and micro F1, respectively. 
    TMa-F stands for type macro F1.
    BERT-contd is incrementally pre-trained in the same setting as ConcEPT except for ECP. 
    %For FIGER and Open Entity, ${\clubsuit}$ indicates results from ~\citet{zhang2019ernie},  
    Results of BERT are taken from ~\citet{zhang2019ernie} except TMA-F. Results of other KEPLMs are taken from their original papers.
    }
    \label{tab:conceptualization_exp}
\end{table*}

% Entity Typing
We first evaluate ConcEPT on entity typing. 
% 什么是Entity Typing
Given an entity mention in a specific context, entity typing is to predict its types, which are also referred to as their concepts~\citep{choi-etal-2018-ultra}.  
% 测试conceptualization knowledge
Hence, this task basically evaluates PLMs' conceptualization ability and taxonomic knowledge. 

\paragraph{\textbf{Datasets.}}
% 数据集
We consider two widely-adopted datasets, Open Entity~\citep{choi-etal-2018-ultra} and FIGER~\citep{AAAI125152}.
% Open
Open Entity is a manually labeled dataset with 9 general concepts.
% FIGER
FIGER is a dataset for fine-grained entity typing, with massive training samples labeled via distant supervision. 
% FIGER 不均衡
However, labels in FIGER are notably imbalanced, 
% 长尾现象
%As is shown in Figure ~\ref{},  
%the sample number of its labels follow the power-law distribution %\polish{that forms a long tail}, 
and the top 5 concepts cover 77.5\% of the data.
% 模型表现主要取决于general concepts. 
Hence, the results are mainly influenced by several general and popular concepts, instead of the fine-grained and unpopular ones. 
% 因此，FIGERFine
Therefore, 
to better explore models' performance on fine-grained concepts, 
% 构造了子集
we collect a subset named FIGER-finer 
% 如何构造
by removing concepts that appear more than 10,000 times in the training set, and discarding samples with no labels left. 
We move several samples from the dev set to the test set to keep the latter's scale.

%Top 5 Samples 1550032 Ratio 0.775016

%\paragraph{\textbf{Datasets.}} We consider the following datasets: % to evaluate ConcEPT's conceptualization knowledge.
%\begin{enumerate}
    %\item CiC dataset of COPEN~\citep{peng2022copen}. This is a multiple-choice dataset,  %that requires models to distinguish the most related concepts of entities in specific contexts. It  with hard distractors collected from hyponyms or hypernyms of the correct answers.
   % \item FIGER~\citep{AAAI125152} is an entity typing dataset with 113 fine-grained types. It adopts the multi-label classification setting, where each entity mention may have plural types.   
    %\item FIGER-250k is a subset of FIGER. \polish{Although FIGER is large-scale, it is significantly imbalanced and model performance is dominated by popular types. Hence, we collect a balanced subset of FIGER.}
    %\item Open Entity~\citep{choi-etal-2018-ultra} is a small manually labeled dataset for entity typing with 6 types. It is under the multi-class classification setting, and each entity has exactly one type.    
%\end{enumerate}

\paragraph{\textbf{Settings.}}
%We fine-tune the models in downstream tasks. 
We place special tokens before and after entity mentions to mark their positions, and represent each entity mention with its boundary token representations. 
The models are fine-tuned with binary cross-entropy loss. 
For Open Entity, we report precision, recall and micro F1.
For FIGER and FIGER-finer, we report strict accuracy, loose macro F1 and  micro F1 following~\citep{AAAI125152}, and additionally type macro F1 which averages the F1 score of each concept.
%\polish{We apply early stopping on type macro F1 of the dev set for this metric, and on (loose) micro F1 for other metrics}.
Hyperparameters of all our experiments are listed in Table \ref{tab:hyperparams}.

% 我们比了哪些baseline
We compare our model with the following PLMs and KEPLMs: 
BERT-base~\citep{devlin2018bert}, ERNIE~\citep{zhang2019ernie},  KEPLER~\citep{wang-etal-2021-kepler}, KnowBERT~\citep{peters-etal-2019-knowledge} and CoLAKE~\citep{sun2020colake}. 
% 参数量是相同的 公平比较 - 需要确定
All these models are based on BERT-base or RoBERTa-base for fair comparison. %the same architecture and pre-trained parameters of BERT-base for fair comparison. 
% 这些模型中 哪些是需要实体链接的 其实不公平
%\polish{Among these models, ERNIE requires entity linking in downstream datasets.} 
% Our BERT
We also report the results of BERT incrementally pre-trained in the same setting 
as ConcEPT except for ECP to highlight the effectiveness of concept-enhanced pre-training.

The results are demonstrated in Table \ref{tab:conceptualization_exp}, from which we have the following observation: 
(1) Compared with vanilla BERT, ConcEPT gains significant improvement with an absolute increase of 2.8\% and 2.5\% on micro F1 of Open Entity and FIGER respectively.  
This proves that concept-enhanced pre-training effectively promotes PLMs's conceptualization ability and taxonomic knowledge.
(2) ConcEPT outperforms existing KEPLMs on these datasets.  
Although CoLAKE barely underperforms ConcEPT on Open Entity, it demands additional entity annotation for downstream datasets, while ConcEPT does not. 
(3) 
%\polish{For ablation study, by comparing among BERT, BERT-contd and ConcEPT, we clearly show that the advancement of ConcEPT originates largely from concept-enhanced pre-training, while continued pre-training are beneficial as well.}
For ablation study, the comparison between ConcEPT and BERT-contd shows that ECP effectively advances conceptual knowledge of PLMs and improve the performance.
%the comparison between ConcEPT and BERT-contd clearly validates the effectiveness of ECP.
(4) On FIGER, the improvement of ConcEPT over BERT is more significant in terms of type macro F1, compared with loose macro and micro F1. 
This indicates that ConcEPT better learns many fine-grained concepts with fewer samples, which type macro F1 concerns more about.
%This is because results of the latter considerably depend on the popular and general concepts in FIGER, which provides abundant samples and vanilla BERT learns well. By contrast, the former concerns more about the fine-grained concepts with fewer samples, which ConcEPT knows better owing with concept-enhanced pre-training. 

\subsection{Conceptual Knowledge Probing}
% 为了进一步探测ConcEPT的概念知识，将其用于COPEN
To further study conceptual knowledge in ConcEPT, we apply it to COPEN~\citep{peng2022copen}, a benchmark of three tasks including conceptual similarity judgment (CSJ), conceptual property judgment (CPJ) and conceptualization in contexts (CiC).
% 具体展开三个任务
CSJ probes whether PLMs recognize similar entities with shared concepts.
CPJ studies PLMs' knowledge of concepts' typical properties. 
CiC requires PLMs to tell the most appropriate concept given an entity and its context.
CSJ and CiC probe PLMs' taxonomic knowledge via multiple-choice classification.
CPJ probes PLMs' concept property knowledge via binary classification. 
%, which also probes PLMs' conceptualization ability and is an extension of entity typing.
% 评价指标，FT / ZP 
We report accuracy of these tasks in both fine-tuning and zero-shot probing settings. 
%The hyperparameters are listed in Table \ref{tab:hyperparams}.
We compare ConcEPT with existing PLMs of different scales. 
Other settings follow ~\citet{peng2022copen}.

The results are shown in Table ~\ref{tab:copen}.
Compared with BERT, ConcEPT achieves improved performance on all tasks and  settings, showing that ECP effectively promotes PLMs' conceptual knowledge.
Specifically, ConcEPT improves by 2.2\% on CPJ and 2.1\% on CiC after fine-tuned, which indicates its enhanced
concept property knowledge and taxonomic knowledge.
Moreover, ConcEPT gains competitive results compared with larger PLMs like T5, with much lower computational costs.  
%Larger PLMs generally outperform BERT, 
%which suggests that pre-training larger models on bigger corpora also improves PLMs' conceptual knowledge. 

%\polish{Nevertheless, ConcEPT still lags behind larger models like T5-base in CPJ and CiC, which suggests that pre-training larger models on bigger corpora also contributes to advanced conceptual knowledge.}

%ConcEPT also acquires an improved understanding of the concepts themselves, rather than their entities only. 

%Recently, COPEN~\citep{peng2022copen} proposes the Conceptualization in Contexts (CiC) task, that requires models to distinguish the most related concepts of entities in specific contexts.

%For CiC COPEN, we follow ~\citet{peng2022copen} and report results on the dev set. %\polish{~\footnote{Labels of the test sets of COPEN are not released.}}.
%We report accuracy in the zero-shot probing (ZP), linear probing (LP) and few-shot learning (FS) settings for CiC COPEN.

\begin{table}[htbp]
    \small
    \centering
    \resizebox{\columnwidth}{!}{
    \begin{tabular}{ccccccc}
    \Xhline{0.8pt}
        \multirow{2}*{Task} &  %& \multicolumn{6}{c}{\textbf{COPEN}}\\ 
    % \cline{2-7}
         \multicolumn{2}{c}{\textbf{CSJ}} & \multicolumn{2}{c}{\textbf{CPJ}} & \multicolumn{2}{c}{\textbf{CiC}} \\ 
    
    \cmidrule(lr){2-3} \cmidrule(lr){4-5} \cmidrule(lr){6-7}
        & \textbf{FT} & \textbf{ZP} & \textbf{FT} & \textbf{ZP} & \textbf{FT} & \textbf{ZP}
        \\
    \Xhline{0.8pt}
        BERT & 27.3 & 20.3 & 68.1 & 49.4 & 49.5 & 37.6 \\ 
        RoBERTa &  22.3 & 15.5 & 72.0 & 49.2 & 52.6 & 31.4 \\ 
        GPT-2 & 20.1 & 7.9 & 70.4 & 51.5 & \textbf{54.2} & 32.3 \\ 
        BART & 21.0 & 14.4 & 68.2 & 48.7 & 51.3 & 33.6 \\ 
        T5 & \textbf{27.9} & 15.2 & \textbf{72.5} & \textbf{55.9} & 53.2 & \textbf{42.3} \\
        ConcEPT & 27.7 & \textbf{21.5} & 70.3 & 52.8 & 51.6 & 38.6 \\
    \Xhline{0.8pt}
    \end{tabular}}
    \caption{Accuracy of tasks in COPEN(\%).  Results of existing PLMs are from ~\citet{peng2022copen}. FT and ZP denote fine-tuning and zero-shot probing, respectively. }
    \label{tab:copen}
\end{table}

\subsection{Relation Classification and Knowledge Graph Completion}
\label{sec:rcandkgc}
Besides, we evaluate ConcEPT on relation classification and knowledge graph completion datasets to show the benefits of concept-enhanced pre-training on other knowledge-intensive tasks. 

\paragraph{\textbf{Relation classification}}
 is to classify the relationship between two entities mentioned in a given context. 
This a widely-studied task of practical importance, and the underlying concepts play a vital role in this task ~\citep{peng-etal-2020-learning}.
Following ~\citet{zhang2019ernie}, we place special tokens before and after the entity mentions to mark their positions, and take the final representation of  [CLS] token for classification. 
We fine-tune PLMs on the TACRED dataset, and report precision, recall and F1. 
Besides, we conduct experiments with BERT and ConcEPT in the \textbf{Only Mention} setting~\citep{peng2020learning}, where the contexts are discarded and only entity mentions are input into PLMs.

According to the results in Table ~\ref{tab:re_exp}, we observe that (1) ConcEPT gains improved performance compared with BERT, which is probably an effect of enhanced conceptual knowledge on relation classification.  
(2) ConcEPT underperforms KEPLER and KnowBert, which suggests the importance of leveraging fact knowledge of KGs in this task.
(3) Our method also increases F1 by 0.9\% in the Only Mention setting, which further proves that ConcEPT learns the concepts of entities and determines the relationship with underlying concepts.

\begin{table}[htbp]
    %\small
    \centering
    \resizebox{\columnwidth}{!}{
    \begin{tabular}{ccccccc}
    \Xhline{0.8pt}
        %\multirow{2}*{Dataset} &
        %\multicolumn{3}{c|}{\textbf{FewRel}} & 
        %\multicolumn{3}{c|}{\textbf{TACRED}} 
        \multirow{3}*{Dataset} &
        \multicolumn{6}{c}{\textbf{TACRED}} 
        \\
    % \cline{2-7}
        & \multicolumn{3}{c}{\textbf{Context \& Mention}} & \multicolumn{3}{c}{\textbf{Only Mention}} \\
    % \cline{2-7}
    \cmidrule(lr){2-4}\cmidrule(lr){5-7}
         & 
        \textbf{P} & \textbf{R} & \textbf{F1} &
        \textbf{P} & \textbf{R} & \textbf{F1} \\
    \Xhline{0.8pt}
        BERT %& 85.1 & 85.1 & 84.9 %85.05 & 85.11 & 84.89 
        & 67.2 & 64.8 & 66.0
        & 53.3 & 40.5 & 46.0\\
        ERNIE %& 88.5 & 88.4 & 88.3 
        & 70.0 & 66.1 & 68.0
        & - & - & -\\
        %RoBERTa %& 85.4 & 85.4 & 85.3
        %& 70.4 & 71.1 & 70.7
        %& - & - & -\\
        KEPLER %& - & - & - 
        & 71.5 & \textbf{72.5} & \textbf{72.0}
        & - & - & -\\
        KnowBert %& - & - & - 
        & 71.6 & 71.4 & 71.5
        & - & - & -\\
        %CoLAKE %& 90.6 & 90.6 & 90.5 
        %& - & - & -\\ \hline
        %Our BERT & 0 & 0 & 0 
        %& 0 & 0 & 0\\ \hline
        ConcEPT %& 86.8 & 86.8 & 86.8 
        & \textbf{74.6} & 64.2 & 69.0
        & \textbf{53.7} & \textbf{41.6} & \textbf{46.9}\\
    \Xhline{0.8pt}
    \end{tabular}}
    \caption{Relation classification results on TACRED (\%).  
    %${\clubsuit}$ and ${\spadesuit}$ indicate results from ~\citet{zhang2019ernie} and ~\citet{wang-etal-2021-kepler} respectively.
    For the Context \& Mention setting, results of BERT  are taken from ~\citet{zhang2019ernie},  and results of other KEPLMs are from the original papers. 
    % 0118 还得压一压 Done?
    } 
    \label{tab:re_exp}
\end{table}

\paragraph{\textbf{Knowledge graph completion}}
 refers to two important tasks for KG construction and representation,  triple classification (TC) and link prediction (LP).
TC is to judge whether a triple  is true or not. 
LP is to predict the missing entity in a corrupted triple (?, r, t) or (h, r, ?), where ? denotes an entity removed for prediction. Typically, the models need to yield a ranking list of the entity set, and the metrics are based on ranks of the correct entities.
We use FB15k-237 as our dataset, and report accuracy for TC, and filtered mean reciprocal rank (MRR) and Hits\@10 for LP. We report averaged results of three runs. 
While previous PLM-based methods like KG-BERT~\citep{kgbert} largely leverage entities' textual descriptions for practical performance, 
our focus is to probe knowledge in PLMs, so we consider a low-resource setting where only names of entities are used. 
For TC, we concatenate names of the head entity, relation, and tail entity of each triple as the input, separated by [SEP] tokens. 
We take the final representation of [CLS] token for classification. 
Negative samples are generated by randomly replacing the head or tail entity of each triple.
For LP, we adopt the contrastive learning framework in LMKE~\citep{wang2022language}.
% 还有entity token，太细节了就算了

The results are demonstrated in Table ~\ref{tab:KGC}, which shows that ConcEPT surpasses BERT on both TC and LP. 
Hence, concept-enhanced pre-training can also benefit knowledge graph completion. 

\begin{table*}
\small
\setlength\tabcolsep{3pt}
\centering
\resizebox{\textwidth}{!}{
\begin{tabular}{lccccccc}
% \hline
    \Xhline{0.8pt}
    %\makecell[c]{\textbf{Case Study}} \\
    % \multicolumn{8}{|c|}{\textbf{Case Study}} \\
    %\hdashline
    % \hline
    %\textbf{Entity Typing}\\
    
    \multirow{2}{*}{\textbf{Samples from FIGER}} & \multirow{2}*{\textbf{Label}} & \multicolumn{3}{c}{BERT} & \multicolumn{3}{c}{ConcEPT} \\
    \cmidrule(lr){3-5} \cmidrule(lr){6-8}
    & & Score & & Similarity & Score &&  Similarity \\ \Xhline{0.8pt}
    
    \multirow{2}*{\textcolor{blue}{\textbf{Hopkins}} said four fellow elections is curious, considering the size of his faculty.}& 
    \multirow{2}*{person} & \multirow{2}*{0.24}& \multirow{2}*{\xmark} & \multirow{2}*{0.53} & \multirow{2}*{0.97}& \multirow{2}*{\cmark} & \multirow{2}*{0.61} \\
    &&&&&&& \\ \cline{1-8}
    
    The building was a partnership between the city and the \textcolor{blue}{\textbf{Spurs}}, ..., and the team&
    \multirow{2}*{sports team} & \multirow{2}*{0.38}& \multirow{2}{*}{\xmark} & \multirow{2}*{0.46} & \multirow{2}*{0.95}& \multirow{2}*{\cmark} & \multirow{2}*{0.50} \\
    contributing the rest.  &&&&&&& \\ \cline{1-8}
    
    The ``working '' German shepherds he cared for were attack dogs or narcotics &
    \multirow{2}*{hospital} & \multirow{2}*{0.02}& \multirow{2}*{\xmark} & \multirow{2}*{0.60} & \multirow{2}*{0.64}& \multirow{2}*{\cmark} & \multirow{2}*{0.69} \\
    sniffers, he said , standing in his year-old \textcolor{blue}{\textbf{Baxter Creek Veterinary Clinic}}.&&&&&&& \\ \cline{1-8}
    
    \multirow{2}*{While serving he taught himself \textcolor{blue}{\textbf{acupuncture}}.}  & 
    \multirow{2}*{treatment} & \multirow{2}*{0.39}& \multirow{2}*{\xmark} & \multirow{2}*{0.58} & \multirow{2}*{0.83}& \multirow{2}*{\cmark} & \multirow{2}*{0.63} \\
    &&&&&&& \\  
    
    % \cline{1-1}
    \Xhline{0.8pt}
    \multirow{2}*{\textbf{Samples from Open Entity}} & \multicolumn{7}{c}{} \\
    & \multicolumn{7}{c}{} \\ 
    \Xhline{0.8pt}    
    % \cline{1-1}
    
    Other scholars have suggested \textcolor{blue}{\textbf{the landing spot}}  may have been on Samana Cay &
    \multirow{2}*{place} & \multirow{2}*{0.10}& \multirow{2}*{\xmark} & \multirow{2}*{0.57} & \multirow{2}*{0.88}& \multirow{2}*{\cmark} & \multirow{2}*{0.61} \\
    or Plana Cays. &&&&&&& \\ \cline{1-8}
    
    In all , \textcolor{blue}{\textbf{it}} visited 62 German train stations , including weeklong stops at several & \multirow{2}*{organization} & \multirow{2}*{0.47}& \multirow{2}*{\xmark} & \multirow{2}*{0.69} & \multirow{2}*{0.82}& \multirow{2}*{\cmark} & \multirow{2}*{0.78} \\ 
    spots in Berlin , organizers said. &&&&&&& \\ 
    \Xhline{0.8pt}
    
\end{tabular}}
\caption{
Case studies for concept-based knowledge transfer on entity typing datasets. The entities are colored blue. 
The samples are from the test set, and we report their scores and average entity similarity with training samples of the same label, given by BERT and ConcEPT. 
}
\label{tab:case_study}
\end{table*} 

\begin{table}[htbp]
    \centering
    \resizebox{\columnwidth}{!}{
    \begin{tabular}{ccccccc}
    \Xhline{0.8pt}
        \multirow{3}*{Dataset} & \multicolumn{3}{c}{\textbf{FB15k-237}} & \multicolumn{3}{c}{\textbf{Wiki-CKT}}\\ 
    % \cline{2-7}
    % \cmidrule(lr){2-4} \cmidrule(lr){5-7}
        & \textbf{CLS} & \multicolumn{2}{c}{\textbf{Link Prediction}} & \textbf{CLS} & \multicolumn{2}{c}{\textbf{Link Prediction}} \\ 
    % \cline{2-5} 
    \cmidrule(lr){2-2} \cmidrule(lr){3-4} \cmidrule(lr){5-5}\cmidrule(lr){6-7}
        & Acc & MRR & Hits@10 & Acc & MRR & Hits@10
        \\
    % \cmidrule(lr){2-2} \cmidrule(lr){3-4} \cmidrule(lr){5-5}\cmidrule(lr){6-7}
    \Xhline{0.8pt}
        BERT & 96.17 & 25.09 & 40.85 & 
        84.93 & 11.22 & 22.33 \\ 
        & $\pm$ 0.07 & $\pm$ 0.19 & $\pm$ 0.32 
        & $\pm$ 0.42 & $\pm$ 0.10 & $\pm$ 0.22 \\ 
        % \hline
        ConcEPT & \textbf{96.22} & \textbf{25.26} & \textbf{40.96} &
    \textbf{85.43} &\textbf{11.89} & \textbf{23.26} \\
        & $\pm$ 0.01 & $\pm$ 0.14 & $\pm$ 0.12
        & $\pm$ 0.56 & $\pm$ 0.12 & $\pm$ 0.29\\ 
    % \hline
    \Xhline{0.8pt}
    \end{tabular}}
    \caption{Results of triple classification (CLS) and link prediction on  FB15k-237 and Wiki-CKT(\%). }
    \label{tab:KGC}
\end{table}

\subsection{Concept-based Knowledge Transfer}

% 概念间的知识迁移
Concepts are important for humans because they lay the basis for our understanding of new entities by enabling knowledge transfer among their instances (e.g., from \textit{Plato} to \textit{Aristoxenus}). 
% 因此，我们研究ConcEPT是否能进行这样的Knowledge Transfer
We study whether ConcEPT is capable of such knowledge transfer with experimental results and cases.

% 具体：数据集，任务
We first construct a KGC dataset named Wiki-CKT (\textbf{C}oncept-based \textbf{K}nowledge \textbf{T}ransfer) to evaluate such ability of PLMs. 
Specifically, we construct 
% 称之为conceptually related pairs of facts, 10000组
collect 2,000 conceptually related groups of facts from Wikidata. 
% support / query
Each group consists of a support fact and two query facts that differ only in either head or tail entities, and the three differential entities share common concepts. 
% 不在训练集
Differential entities in query facts do not appear in support facts. 
% Long tail 
%\polish{We focus on concepts' long-tail instances ... }
%The query instances set and the support instances set are disjoint.
% Trainset
The support fact is put into the training set, while 
% Valid & Test
the two query facts are split into the dev and test set.  
%We name this dataset Wiki-CKT (\textbf{C}oncept-based \textbf{K}nowledge \textbf{T}ransfer). 
Wiki-CKT contains many long-tail entities. %rarely mentioned in Wikipedia and Wikidata.
%Its statistics are shown in Table ~\ref{tab:statistics}.
We fine-tune and evaluate ConcEPT and BERT-base on both triple classification and link prediction on Wiki-CKT. 
The settings are mostly the same as in Sec \ref{sec:rcandkgc}.
For TC, negative samples are generated by replacing the differential entities.
%The hyperparameters are listed in Table ~\ref{tab:hyperparams}. 

% compare with ?
The results are shown in Table ~\ref{tab:KGC}, and we observe that ConcEPT improves by 1.5\% and 1.0\% over BERT on accuracy of TC and Hits\@10 of LP, which validates ConcEPT's improved capability of concept-based knowledge transfer.
%As we input only names of entities and relations without additional contexts or descriptions, this improvement originates from concept-enhanced representations of entities. 

%\polish{We are also interested in concept to instances, name syllogism reasoning. }
%Link Prediction in Syllogism Setting

Besides, we conduct case studies on FIGER and Open Entity to further demonstrate concept-based knowledge transfer. 
The cases are from the test sets, and we report their scores and average entity representation similarity with training samples of the same label given by BERT and ConcEPT. 
The similarity is calculated before fine-tuning. 
As is shown in Table ~\ref{tab:case_study}, ConcEPT successfully recognizes instances of both general concepts like \textit{person}, \textit{place} and fine-grained concepts like \textit{sports team} and \textit{hospital}. 
Such improvement results partly from concept-based knowledge transfer.
With concept-enhanced pre-training, we observe notably increased entity similarity within the same label (concept), which helps models to better generalize from training samples to test ones with underlying concepts. 
%Hence, after models are fine-tuned on the training set, ConcEPT  better generalize to test samples, as they are more similar to training samples of the same label. 
This applies to both widely-known entities like \textit{Spurs}, long-tail entities like \textit{Baxter Creek Veterinary Clinic}, and pronouns or coreference like \textit{it} and \textit{the landing spot}, which suggests that ConcEPT not only learns concepts of named entities, but also learns concept-aware language representations to infer appropriate concepts from  contexts.

\section{Conclusion}

In this paper, we propose concept-enhanced pre-training to infuse conceptual knowledge into PLMs. 
We introduce entity concept prediction, a novel pre-training objective to predict concepts of mentioned entities in the contexts, which is supported by \textit{isA} relations from external taxonomies. 
Experimental results show that concept-enhanced pre-training effectively improves model performance on various knowledge-intensive tasks, which suggests that our ConcEPT model gains enhanced conceptual knowledge, including taxonomic knowledge and concept property knowledge. 
Additional case studies further show that ConcEPT acquires an improved capability of concept-based knowledge transfer.

\section*{Limitations}
In this section, we discuss some limitations of our work.
% 模型大小和结构
First, the effectiveness of concept-enhanced pretraining on models with different architectures and scales remains to be  studied. 
In this paper, we only experiment with BERT as the base model due to  limited computational resources. 
% 从模型结构上改进
Second, we do not explore enhancing PLMs via model architecture.
This work focuses on using concepts as supervision to infuse conceptual knowledge, which is more convenient for downstream applications. 
However, a taxonomy-aware architecture is potentially also powerful towards improved conceptual knowledge. 
% 概念选择、taxonomy构建
Finally, we do not explore the influence of the selection of concepts and taxonomies for concept-enhanced pre-training. 
Noisy concepts and \textit{isA} relations are common in existing taxonomies, which may have unwanted influence on PLMs. Hence, better methods to select or denoise \textit{isA} relations are expected.

% 未来：相应探索以上三点
Accordingly, in the future, we will (1) explore infusing conceptual knowledge into PLMs of other structures such as text-to-text models, (2) study taxonomy-aware architecture for concept-enhanced models, as well as (3) investigating the influence and strategy for the selection of concepts and taxonomies. 

%to further promote conceptual knowledge of PLMs, especially with taxonomy-aware model architectures. 

%We only enhance pre-training on the output side with concepts, and haven't study enhance PLMs on the input side, which trades off ease of use with possible improvement. 

%Add concept tokens as inputs, downstream tasks, better communiate different samples, help PLMs learn with underlying concepts .

\section*{Ethics Statement}
We discuss the ethical concerns of our work:
(1) Intellectual property and intended use. 
For pre-training, we use data dump of Wikipedia and Wikidata, which are under the CC BYSA 3.0 license~\footnote{https://creativecommons.org/licenses/by-sa/3.0/}.
For data processing, Wiki-extractor is shared under the GNU Affero General Public License v3.0~\footnote{https://www.gnu.org/licenses/agpl-3.0.en.html}, Wikidata-QRank is shared under CC0-1.0~\footnote{https://creativecommons.org/publicdomain/zero/1.0/}, and Tagme is under the CC BY-SA 3.0 and Apache License 2.0~\footnote{https://www.apache.org/licenses/LICENSE-2.0}.
For downstream tasks, COPEN is released under the MIT License~\footnote{https://opensource.org/licenses/MIT}. 
TACRED is released via the Linguistic Data Consortium (LDC) under LDC User Agreement for Non-Members, which cannot be redistributed. 
Other data and codes, are all public and established resources and support extensive academic researches especially in the field of natural language processing.  
We believe these resources are well anonymized to avoid potential risks and do not  contain offensive content.
Our newly created artifacts follow their intended use and access conditions. 
%BERT, huggingface Transformer, its own terms
%FIGER no license, 
%Open Entity no license
%FB15k-237 no license, 
%COPEN MIT License, TACRED
%TACRED is released via the Linguistic Data Consortium (LDC)
%LDC User Agreement for Non-Members
%Our use of existing artifacts follows their intended use and previous works.
%For the artifacts we create, our intended use is consistent with their original access conditions.
(2) Human Annotation. We recruit two annotators from undergraduate students for concept annotation.  
They are adequately paid according to their working hours given their demographic. 
They are well-informed of how the data will be used and released.

% Entries for the entire Anthology, followed by custom entries
\bibliography{anthology,custom}
\bibliographystyle{acl_natbib}

\appendix
\begin{table*}[!htb]
    \centering
    \setlength{\tabcolsep}{0.15cm}{
    \resizebox{\textwidth}{!}{
    \begin{tabular}{*{12}{c}}
    \toprule
    \multirow{2}*{} & \multirow{2}*{Open Entity} & \multirow{2}*{FIGER} & \multirow{2}*{FIGER-finer} &  \multicolumn{3}{c}{COPEN}   &  \multirow{2}*{TACRED} & \multicolumn{2}{c}{FB15k-237} & \multicolumn{2}{c}{Wiki-CKT} \\
    \cmidrule(lr){5-7} \cmidrule(lr){9-10} \cmidrule(lr){11-12} & & & & CSJ & CPJ & CiC  &  & TC & LP & TC & LP\\
     %& Open Entity & FIGER & FIGER-finer & CSJ & CPJ & CiC   &  TACRED & FB15k-237 & Wiki-CKT \\
    
    \midrule
    
     learning rate & 1$e$-5 & 2$e$-5 & 2$e$-5 
     & 3$e$-5 & 3$e$-5 & 2$e$-5 
      & 2$e$-5 & 2$e$-5 & 2$e$-5 & 2$e$-5 & 2$e$-5  \\
     batch size & 16 & 2048 & 2048
     & 8 & 64 & 4 
      & 32 & 128 & 128 & 128 & 128  \\
     epoch & 15 & 5 & 30
     & 5 & 10 & 5 
      & 5 & 5 & 10 & 10 & 30  \\
    max length & 256 & 256 & 256 & 128 & 128 & 120 & 256 & 512 & 512 & 512 & 512 \\
     %weight decay & 1$e$-2 & 1$e$-2 & 1$e$-5 & 0 & 0 & 0 & 0 & 0 & 0 & 0 \\ 
    \bottomrule
    \end{tabular}}}
    \caption{Hyperparameters for downstream datasets in this paper. TC and LP indicate triple classification and link prediction respectively. }
    \label{tab:hyperparams}
\end{table*}

\section{Concept Annotation}
In this section, we describe the details of how we instruct annotators to manually 
process the concepts, aiming at selecting concepts that provide high-quality supervision. 
%which aims at removing noisy and non-informative concepts while merging similar ones. 

We invite two annotators to annotate 1761 concepts after the automatic selection steps. 
The annotators are instructed to annotate each concept as ``select'', ``remove'' or ``merge with a  selected concept''.  
Afterward, one author of this paper deals with inconsistent annotations.  

The annotators are informed of the usage of data, and are asked to select concepts that would provide high-quality supervision to our method. They are informed of the following instructions:

\begin{enumerate}
    \item The concepts should be meaningful and provide knowledge to their entities. For example, concepts like \textit{term}, \textit{taxon} and \textit{genre} provide little knowledge for people to understand their entities, and are hence removed. 
    \item The concepts should be at a basic level, which the annotators can easily categorize their typical entities as. For example, the annotators may categorize \textit{University of California, Davis} as \textit{university} or \textit{school}, instead of \textit{census-designated place}. Complex concepts should be merged or removed. 
    \item Concepts with many noisy entities (noisy \textit{isA} relations) should be removed. For example, the concept \textit{style} is removed because  many of its entities are historical countries and periods like \textit{Tang Dynasty}.
    \item A concept may correspond to multiple Wikidata items, which should be merged. Highly similar concepts should also be merged.
    \item The concepts should be well associated with contexts describing their entities. 
    For example, the concept \textit{farmer} has popular entities like \textit{George Washington} and \textit{Abraham Lincoln}. They do not represent typical farmers and their contexts are mainly related to \textit{politician}. 
    Also, a concept's entities should be mentioned in relatively similar contexts. For example, the concept \textit{technique} contains entities \textit{electronics} and \textit{bronze sculpture}, whose contexts are largely different.
\end{enumerate}

Overall, we obtain 1305 concepts, with 386 concepts removed and 70 concepts merged into other concepts. 

\section{Hyperparameters}
\label{sec:appendix}

We list the hyperparameters used to fine-tune our models on downstream datasets in Table \ref{tab:hyperparams}.

\end{document}